\pdfoutput=1

\documentclass[11pt]{article}

\usepackage[]{ACL2023}

\usepackage{times}
\usepackage{latexsym}
\usepackage{xcolor}

\usepackage[T1]{fontenc}

\usepackage[utf8]{inputenc}

\usepackage{microtype}

\usepackage{inconsolata}

\title{How Ready are Pre-trained Abstractive Models and LLMs for Legal Case Judgement Summarization?}

\author{Aniket Deroy \\
  IIT Kharagpur \\
  West Bengal 721302, India \\
  \texttt{roydanik18@gmail.com} \\\And
  Kripabandhu Ghosh \\
  IISER Kolkata \\
  West Bengal 741246, India \\
  \texttt{kripa.ghosh@gmail.com} \\ \And
  Saptarshi Ghosh \\
  IIT Kharagpur \\
  West Bengal 721302, India \\
  \texttt{saptarshi@cse.iitkgp.ac.in} \\ 
}

\begin{document}

\maketitle

\begin{abstract}
Automatic summarization of legal case judgements has traditionally been attempted by using extractive summarization methods. However, in recent years, abstractive summarization models are gaining popularity since they can generate more natural and coherent summaries. Legal domain-specific pre-trained abstractive summarization models are now available. 
Moreover, general-domain pre-trained Large Language Models (LLMs), such as ChatGPT, are known to generate high-quality text and have the capacity for text summarization. Hence it is natural to ask if these models are ready for off-the-shelf application to automatically generate abstractive summaries for case judgements. To explore this question, we apply several state-of-the-art domain-specific abstractive summarization models and general-domain LLMs on Indian court case judgements, and check the quality of the generated summaries. In addition to standard metrics for summary quality, we check for inconsistencies and hallucinations in the summaries. 
We see that abstractive summarization models generally achieve slightly higher scores than extractive models in terms of standard summary evaluation metrics such as ROUGE and BLEU. However, we often find inconsistent or hallucinated information in the generated abstractive summaries. 
Overall, our investigation indicates that the pre-trained abstractive summarization models and LLMs are not yet ready for fully automatic deployment for case judgement summarization; rather a human-in-the-loop approach including manual checks for inconsistencies is more suitable at present.
\end{abstract}


\section{Introduction}

Summarization of legal case judgements  is a practical and important problem in the legal domain, given that the extreme length and complexity of such documents make it difficult even for Law practitioners to read them fully. 
Traditionally, case judgements are summarized by humans, i.e., Law practitioners. For instance, most Legal information systems provide case summaries/headnotes written by Law practitioners. 
To reduce the human effort in summarization, there have been many efforts over the years to automate the summarization of case judgements~\cite{bhattacharya2021incorporating,deroy2023ensemble}.

There are two broad approaches for summarization - Extractive (where some important sentences are selected from the input document to form the summary) and Abstractive (where the model attempts to understand the document and generate a summary on its own).
The reader is referred to the comprehensive surveys by Nenkova et al.~\cite{Nenkova2012ASO} and Wafaa et al.~\cite{summarization-survey-wafaa} for more details on various types of summarisation algorithms.

For summarization of legal case judgements, extractive summarization models have mostly been applied over the years~\cite{bhattacharya2021incorporating,polsley-etal-2016-casesummarizer,liu2019extracting,10.1145/3322640.3326728}. 
But in recent times, the research community is preferring the use of {\it abstractive} summarization models, primarily because abstractive methods are said to generate more `natural' and `coherent' summaries.
As a result, a few recent works have started training abstractive models for legal document summarization~\cite{shukla2022legal,feijo2023improving}.
Domain-specific pre-trained versions of popular abstractive summarization models, such as Google's Pegasus~\cite{pegasus}, have been released specifically for legal summarization (e.g., Legal Pegasus -- \url{https://huggingface.co/nsi319/legal-pegasus}).
Moreover, recent times have seen the advent of general-purpose Large Language Models (LLMs) such as ChatGPT and DaVinci that have the ability to generate high-quality text as well as the ability to summarize text without additional training. 
A big advantage of these {\it pre-trained} abstractive summarization models and LLMs is that they can be applied without further training.
In fact, LLMs are already being used for summarization in other domains, e.g., news summarization~\cite{zhang2023benchmarking}. 
But, to our knowledge, these LLMs have not been much used for legal case judgement summarization to date. 

In such a scenario, it is natural to ask -- \textit{how ready are the pre-trained abstractive summarization models and the LLMs that are available today, for off-the-shelf application for legal case judgment summarization?}
In this paper, we attempt to answer this question.

We apply state-of-the-art abstractive summarization models specifically meant for the legal domain -- such as Legal-Pegasus (\url{https://huggingface.co/nsi319/legal-pegasus}) and Legal-LED (\url{https://huggingface.co/nsi319/legal-led-base-16384}) -- as well as recently developed Large Language Models such as DaVinci and ChatGPT, on a dataset of Indian Supreme Court case judgements (containing gold standard summaries written by Law practitioners). 
We also apply some extractive summarization models on the same dataset for comparison.
We report a large number of summary quality metrics for all the models, including traditional metrics such as ROUGE, METEOR and BLEU (that match model-generated summaries with gold standard summaries) and metrics for quantifying the consistency of summaries with respect to the original document. 

We observe that the summaries generated by abstractive models achieve slightly higher ROUGE, METEOR, BLEU scores than those generated by the extractive models. 
However, the abstractive summaries have various problems, including incomplete sentences/words, multiple sentences being merged meaninglessly, as well as more serious errors such as inconsistent and hallucinated information. 
For instance, we observe that the abstractive summarization models and LLMs sometimes generate wrong dates and wrong person names in the summaries, and also confuse different persons associated with a case. 
Thus our contributions in this work are as follows:\\
(1) We apply pre-trained abstractive summarization models and LLMs (and a few extractive summarization models for comparison) on a set of Indian court case judgements, and report several metrics that include not only traditional summarization evaluation metrics, but also metrics for the consistency of the generated summaries.\\ 
(2) To our knowledge, this paper is the first analysis of the consistency of abstractive summaries in the legal domain. We show that, though abstractive models often achieve higher ROUGE, BLEU, METEOR scores than extractive models, abstractive summaries often contain  hallucinated or inconsistent information.\\
(3) We present several examples of errors, including presence of hallucinated or inconsistent information, in case judgement summaries generated by state-of-the-art LLMs and pre-trained abstractive summarization models. To our knowledge, this is the first study to demonstrate such examples.
 
Our analyses show that the pre-trained abstractive summarization models and LLMs need to be further improved before they can be readily used for case judgement summarization by legal experts.


\section{Related work}

\noindent \textbf{Summarization of legal case judgements:}
Traditionally, \textit{extractive} summarization models have been used to summarize legal case judgements. A variety of methods have been tried including optimization techniques~\cite{bhattacharya2021incorporating}, multi-task learning~\cite{agarwal2022extractive}, Machine Learning-based classification~\cite{liu2019extracting}, and so on. The extractive models that have been tried include both unsupervised~\cite{bhattacharya2021incorporating} and supervised~\cite{agarwal2022extractive,liu2019extracting} models.


In recent times, there have been a few works on \textit{abstractive} summarization of legal case judgements. 
Our recent prior work~\cite{shukla2022legal} applied various abstractive models such as BART, Legal-LED and Legal-Pegasus on Indian and UK court judgements. 
There are prior works on semantic segmentation of long legal documents in low resource settings, which discuss how to handle long legal documents (which are generally larger than the input length of encoder-decoder based models) to perform abstractive legal document summarization~\cite{moro2022semantic}.
There are works which try to improve abstractive summarization of legal case judgements using textual entailment~\cite{feijo2023improving}.



\vspace{3mm}
\noindent \textbf{Hallucinations in large language models:}
In the context of natural language processing (NLP), hallucination refers to a phenomenon where a language model generates text that is not true or accurate based on the input it has been given. 
This can happen for a variety of reasons, such as a lack of training data, bias in the training data, or limitations in the language model architecture (see~\cite{ji2023survey} for a survey).

There have been studies on hallucination specifically in abstractive summaries. Since hallucinations are undesirable in summaries, various works have tried to reduce hallucinations in the summaries generated by the abstractive summarization models~\cite{filippova-2020-controlled,zhao2020reducing}. 


The advent of Large Language Models (LLMs) like ChatGPT, and their increased use in academic writing is raising further concerns about the integrity and accuracy of the generated text~\cite{alkaissi2023artificial}. While such models are trained on vast amounts of data and can produce high-quality content, there is always a risk that the generated text may contain inaccuracies, biases, or even outright fabrications.
For example, language models trained on Wikipedia and other online sources have been found to generate more sexist and racist content~\cite{stanczak2021survey}.
Additionally, LLMs can also generate text that is inconsistent with established scientific facts or that presents misleading information.



\vspace{3mm}
\noindent \textbf{Novelty of this work:} There has been little attempt to analyse how various {\it abstractive} summarization methods and LLMs (such as ChatGPT) perform in summarizing legal case judgements. 
Also, to our knowledge, hallucination  has not been studied earlier in the context of legal summarization.
This work takes the first step towards understanding how prepared the abstractive summarization models / LLMs are today for the task of automatic case judgement summarization.


\section{Dataset}
\label{sec:dataset}


We reuse a dataset of Indian Supreme Court judgements from our prior work~\cite{shukla2022legal}.
The dataset, called IN-Abs, contains a total of 7,130 legal judgements from the website of the Legal Information Institute of
India\footnote{\url{http://www.liiofindia.org/in/cases/cen/INSC/}}, 
along with a single abstractive summary for every judgement. The summaries (also known as `headnotes') have been written by Law experts appointed by Legal Information Institute of India. 

\begin{table}[tb]
\centering
\small
\begin{tabular}{|p{0.15\columnwidth}|p{0.15\columnwidth}|p{0.2\columnwidth}|p{0.23\columnwidth}|}
\hline
\textbf{} & Nos. of Documents & Avg nos. of words in document & Avg nos. of words in gold-standard summary 
\\ \hline 
Train set & 7,030 & 4,368.49 & 839.75 \\ \hline
Test set & 100 & 4,782.71 & 932.01 \\ \hline

\end{tabular}
\caption{Statistics of the IN-Abs train set and test set, containing (case judgement, summary) pairs from the Indian Supreme Court. The train set is used to train extractive models and fine-tune pre-trained abstractive models. All summarization models in this work are applied and evaluated over the test set.}
\label{tab:IN-Abs-stats}
\end{table}


Out of the total set of 7,130 judgement-summary pairs in the dataset, 7,030 judgement-summary pairs are considered as the training set and the other 100 judgements are considered as the test set.
Some of the supervised abstractive/extractive models considered in this work have been trained or fine-tuned over the IN-Abs train set. 
All summarization models are evaluated over the IN-Abs test set (100 documents).

Table~\ref{tab:IN-Abs-stats} represents the number of documents in the training and test sets, along with the average number of words present in a legal judgement and a gold standard summary. Further details about the IN-Abs dataset are available in~\cite{shukla2022legal}.




\section{Methods for summarizing legal case judgements}
\label{sec:methods}

We have tried a variety of summarization models in this work. 
There are 3 main categories of summarization methods applied in this work: 
(1)~General-domain Large Language models, 
(2)~Legal domain-specific abstractive summarization models, and 
(3)~Extractive Summarization models.

\subsection{General-domain Large Language Models}
\label{sec:legal_led}

We try out two popular Large language Models (LLMs), namely, Text-Davinci-003 and Turbo-Gpt-3.5, both developed by OpenAI.\footnote{Details of the two LLMs are available at \url{https://platform.openai.com/docs/models/}.}

\textbf{Text-Davinci-003} (which we refer to as \textbf{Davinci} in short) is a transformer-based language model with 175 billion parameters, making it one of the largest and most advanced language models to date. The language model has been trained on a diverse range of text data, including web pages, books, scientific articles, and other sources of human-written text. OpenAI has not provided detailed information on the exact sources of the training data, but it is known that the model has been trained on a massive scale text dataset using a combination of supervised and unsupervised learning methods.

\textbf{Turbo-GPT-3.5} (popularly known as \textbf{ChatGPT}) is a language model which is based on the GPT-3 architecture developed by OpenAI. The model is said to have approximately 154 billion parameters. 
Turbo-GPT-3.5 was trained on a diverse range of text data, including web pages, books, scientific articles, and other sources of human-written text including \textit{chats}, using a combination of supervised and reinforcement learning methods. The model has been optimized for speed and performance, with efficient use of memory and computation resources.

Davinci is said to be the largest and most powerful model till date, which performs the best on many complex NLP tasks. ChatGPT is a cheaper model with slightly fewer parameters; though it is said to be `optimized for chat', ChatGPT also performs very well in many types of NLP tasks. 

Both these LLMs take as input a `prompt' and generate text in response. Specifically for the summarization task, the prompt consists of (i)~the text to be summarized, which we refer to as \verb|<text to summarize>| and (ii)~an `instruction' that tells the model that the input text has to be summarized.
For both the LLMs -- Text-Davinci-003 and Turbo-GPT-3.5 -- we consider two variations giving two different prompts for summarization, as explained below. 


\vspace{2mm}
\noindent \textbf{Variations of Text-Davinci-003:} We try these two variations of the model:-\\
(i)~\textbf{davinci-tldr}: for this model, the prompt is ``\verb|<text to summarize> Tl;Dr|''. In other words, the text to be summarized is passed first followed by  ``Tl;Dr'' which is an inbuilt identifier for summarization.\footnote{\url{https://platform.openai.com/examples/default-tldr-summary}}\\
(ii)~\textbf{davinci-summ}: for this model, the prompt is ``\verb|<text to summarize>| Summarize the document in \verb|<XX>| words'' where XX is a number representing the target length of the output summary in number of words, i.e., the maximum number of words in the summary to be generated. 
How the target length XX is decided will be explained below.

\vspace{2mm}
\noindent \textbf{Variations of Turbo-Gpt-3.5 (ChatGPT):} Similar to what we did for the Davinci model, we try the following two variations:-\\
(i)~\textbf{chatgpt-tldr}: here the prompt is ``\verb|Tl;Dr <text to summarize>|''. In other words, the inbuilt identifier for summarization ``Tl;Dr'' is sent first, followed by the text to summarize.\\
(ii)~\textbf{chatgpt-summ}: for this model, the prompt is ``Summarize the document in \verb|<XX>| words \verb|<text to summarize>|''   where XX is a number representing the target length of the output summary (in words). 
The choice of the target length is discussed below.

\vspace{3mm}
\noindent {\bf Chunking of long legal documents:} \label{sec:chunking}
LLMs such as ChatGPT and DaVinci impose restrictions over the length of input that can be given at once.
In particular, Text-Davinci-003 and Turbo-GPT-3.5 have a \textit{limit of 4,096 tokens for (Prompt + generated text)}, where every `token' represents approx. 4 characters.
On average, one token corresponds to $\frac{3}{4}$ of an English word, or 100 tokens approximately corresponds to 75 words.\footnote{Tokens are explained in detail at \url{https://help.openai.com/en/articles/4936856-what-are-tokens-and-how-to-count-them}.}

Since most legal case judgements are longer than this limit (having more than 4,300 words on average), we have to follow a divide and conquer strategy to summarize long legal documents using these LLMs. 
Given the limit of 4,096 tokens for (Prompt + generated text), we choose to send at most 1,024 words as the text to be summarized (as part of the prompt, as described above) at a time to these LLMs.
Thus, we chunk the legal documents of length higher than 1,024 words and then pass the chunks (one at a time) into Turbo-Gpt-3.5 / Text-Davinci-003 to obtain the output summaries for the chunks. 
The summary for every chunk (of size 1,024 or less) is obtained from these models and then the summaries of all chunks are appended together (in the same order as of the chunks) to form the final output summary for the case judgement document.
For legal documents with length less than 1,024 words, the entire document is passed into the model at once, to obtain the summary.

Note that the performance of summarization models may depend on the size of chunks. 
We conducted experiments with a subset of the documents considering two chunk sizes -- 1,024 words and 2,048 words. 
We observed ChatGPT to perform slightly better with 1,024-word chunks, as per all the summarization evaluation metrics (the metrics will be detailed in the next section). 
Whereas, Davinci gave slightly better values for a few of the metrics with 1,024-word chunks, and better values for the other metrics with 2,048-word chunks. 
For simplicity and consistency, in this work, we report all results considering chunks of size at most 1,024 words for all models. Further exploration of the dependence of summarization performance on the chunk size is left as future work.

\vspace{3mm}
\noindent {\bf Deciding the target summary length for a chunk:}
When some text is sent to a LLM for summarization, we need to specify the target summary length in the `max tokens' hyperparameter, i.e., the maximum number of words in the summary to be generated. 

Suppose a chunk of text of length 1024 words from a document $D$ is sent to a LLM for summarization. Let the length of document $D$ be $|D|$ words, and the length of the gold standard summary of $D$ be $|S|$ words.
Then the target summary length for the chunk is specified as $\frac{|S|}{|D|} \times 1024$ words.
In other words, we ask the LLM to summarize each chunk considering the same compression ratio as for the whole document and the gold standard summary. 

There is an inherent limitation in this method, which is as follows. In reality, all parts of the document are \textit{not} equally important, hence different chunks should possibly be allocated different lengths in the final summary. In contrast, this method allocates the same length in the summary for all chunks.
However, there is no simple way of knowing the relative importance of different chunks in a legal case judgement.

\begin{table*}[tb]
\centering
\small
\begin{tabular}{|p{0.15\textwidth}|p{0.75\textwidth}|}
\hline
\textbf{Model} & \textbf{Hyperparameters} \\ \hline
chatgpt-tldr & temperature=0.7 , max tokens = gold-std summary length * 1024/Document length. \\ \hline

chatgpt-summ & temperature=0.7 , max tokens = gold-std summary length * 1024/Document length. \\ \hline

davinci-tldr & Presence penalty=1.0,  frequency penalty=0.0, temperature=0.7, \newline max tokens = gold-std summary length * 1024/Document length.   \\ \hline

davinci-summ & Presence penalty=1.0,  frequency penalty = 0.0, temperature=0.7, \newline
max tokens = gold-std summary length * 1024/Document length.   \\ \hline

LegPegasus & max tokens = gold-std summary length * 1024/Document length.    \\ \hline

LegPegasus-IN & max tokens = gold-std summary length * 1024/Document length.    \\ \hline

LegLED & max tokens = gold-std summary length * 1024/Document length.  \\ \hline

LegLED-IN & max tokens = gold-std summary length * 1024/Document length. \\ \hline

\end{tabular}
\caption{Hyperparameters of the legal domain-specific abstractive models and LLMs used in the work. `max tokens' indicates the maximum number of words in the summary to be generated for an input chunk of text of length 1,024 words. Here `gold-std summary length' is the actual length (number of words) of the gold standard summary for the given document.}
\label{tab:hyperparameters}
\end{table*}

\vspace{2mm}
\noindent \textbf{Implementation details:} 
The LLMs stated above have been run using the OpenAI API\footnote{\url{https://platform.openai.com/docs/api-reference/completions}}.
The hyperparameters of Text-Davinci-003 and Turbo-GPT-3.5 are indicated in Table~\ref{tab:hyperparameters}. 
We use the default values for the hyperparameters `presence penalty', `frequency penalty' and `temperature'.
The `max tokens' hyperparameter indicates the maximum number of words in the summary to be generated for an input chunk of text; it is computed as described above. 

\subsection{Legal domain-specific abstractive summarization models}
\label{sec:legal_specific}

While the LLMs described in the previous section are general-domain (not trained for any particular domain or task), we now consider some abstractive summarization models that are specifically designed for summarization in the legal domain.

One such model is \textbf{Legal-Pegasus} (which we abbreviate to \textbf{LegPegasus}). This model is based on the google$/$pegasus-cnn\_dailymail model developed by Google, which is designed to perform abstractive summarization task.
LegPegasus has been specifically designed for the legal domain by finetuning it on 
the `sec-litigation-releases' dataset consisting of more than 2,700 litigation releases and complaints concerning civil lawsuits in various courts in the USA (and their summaries) brought by the US Securities and Exchange Commission. 
The LegPegasus model is available at \url{https://huggingface.co/nsi319/legal-pegasus} and has a maximum input sequence length of 1024 tokens.

Another abstractive summarization model specifically designed for the legal domain is \textbf{Legal-LED} (Legal Longformer Encoder Decoder) which we abbreviate as \textbf{LegLED}.
The LegLED model is based on the Longformer architecture, a transformer-based neural network architecture that has been specifically designed for processing long sequences of text. 
The LegLED, available at \url{https://huggingface.co/nsi319/legal-led-base-16384}, has been finetuned on the same  `sec-litigation-releases' dataset as described above, to make it suitable for summarization in the legal domain. 

As stated above, both LegPegasus and LegLED have been finetuned over legal documents and their summaries from the US Courts of Law.
To make the models more suitable for summarizing Indian legal documents, our prior work~\cite{shukla2022legal} further finetuned the models over the IN-Abs training set (containing 7,030 Indian case judgements and their summaries, as stated in Section~\ref{sec:dataset}). We call these models \textbf{LegPegasus-IN} and \textbf{LegLED-IN} since they have been specifically finetuned for summarizing Indian legal documents.


\vspace{3mm}
\noindent {\bf Chunking of long legal documents:} Since the domain-specific abstractive models also have restrictions of the number of input tokens, we follow a similar chunking-based strategy to handle long legal documents, as was described in Section~\ref{sec:chunking}. 
We chunk the legal documents (of length higher than 1,024 words) into chunks of at most 1,024 words and then pass one chunk at a time into the summarization models. 
The summary for every chunk is obtained from these models and then appended together (in the same order as the chunks in the source document) to form the final output summary. The target summary length of each chunk is decided as described in Section~\ref{sec:chunking}.
For documents shorter than 1,024 words, the entire summary of the document is obtained at once.





\subsection{Extractive summarization models}

We consider some extractive summarization models for comparison with the abstractive models and LLMs. 
In our prior works~\cite{deroy2023ensemble,shukla2022legal}, we applied several extractive summarization methods on the IN-Abs dataset. 
We observed that the three methods (i)~CaseSummarizer, (ii)~BertSum, and (iii)~SummaRunner/RNN\_RNN performed perform well over the IN-Abs dataset across most metrics.
So we include the following three extractive methods in the comparison.

\vspace{1mm}
\noindent 
\textbf{(1) Case Summarizer}~\cite{polsley-etal-2016-casesummarizer} is an unsupervised method that identifies the most relevant sentences or phrases of a legal case document based on a metric like TF-IDF. CaseSummarizer adjusts sentence scores using occurrences
of known entities, dates, and proximity to section headings. 

\vspace{1mm}
\noindent 
\textbf{(2) BertSum}~\cite{liu2019fine} is a supervised summarization model that uses the Bidirectional Encoder Representations from Transformers (BERT) architecture. 
This model treats summarization as a binary classification problem where every sentence (in the document) is labeled as 1 if the sentence is suitable for inclusion in the summary, and 0 otherwise. 
The model is trained (over a training set containing documents and gold standard summaries) to identify sentences that are suitable for inclusion in the summary.

\vspace{1mm}
\noindent 
\textbf{(3) SummaRunner/RNN\_RNN}~\cite{nallapati2017summarunner} is a supervised model that attempts to identify the most important sentences in a text and generate a concise summary. Similar to BertSum, this model considers summarization as a classification problem, and also analyzes the relationships between sentences in a document to select those that contain the most relevant information.


\vspace{1mm}
\noindent 
For all the three extractive models stated earlier, we use the implementations made available in our prior work~\cite{shukla2022legal}. The supervised models
BertSum and SummaRunner/RNN\_RNN models have been trained on the 7,030 (legal document, summary) pairs in the IN-Abs train dataset.
More details about the training procedure are available in~\cite{shukla2022legal}.




\section{Comparing performances of summarization models}

In the previous section, we described several summarization models, including LLMs, domain-specific abstractive models, and extractive models.
We now compare the quality of summaries generated by the different methods along two aspects -- 
(1)~their match with the gold standard summaries, and 
(2)~their consistency with the input documents.

\subsection{Match with gold standard summaries}

We first discuss the metrics used for measuring the match with gold standard summary, and then compare the performances of the different summarization models according to those metrics.

\subsubsection{Metrics}

We use the following well-known metrics that compare a model-generated summary with the gold-standard summary (written by domain experts) and give a score, where higher scores imply higher match with the gold-standard (and hence a better quality summary). 

\textbf{(1) ROUGE}~\cite{lin2004rouge} (Recall Oriented Understudy of Gisting Evaluation) is possibly the most popular metric used for measuring the quality of a summary generated by a summarization model. In particular, we calculate \textit{Rouge-2} precision, recall and F1 scores that measure the bigram match between gold standard summaries and model-generated summaries, and \textit{Rouge-L} precision, recall and F1 scores which measures Longest Common Subsequence-based match between generated summaries and the gold standard summaries. 

\textbf{(2) METEOR}~\cite{banerjee2005meteor} calculates the harmonic mean of unigram precision and recall and is generally used for evaluating machine translation output.
Prior works have also used this metric to evaluate summaries~\cite{deroy2023ensemble}. 
Here we use this metric to calculate the unigram overlap between a model-generated summary and the gold standard summary. 

\textbf{(3) BLEU}~\cite{papineni2002bleu} (Bilingual Evaluation Understudy) is a metric generally used for evaluating machine translation output, but it can also be used for measuring how well a model-generated summary matches with a gold standard summary.  

\vspace{2mm}
\noindent For all the above metrics, we use the implementations from the SummEval package (\url{https://github.com/Yale-LILY/SummEval}) which is a well-known package for evaluation of summarization.

\begin{table*}[tb]
\centering
\small
\begin{tabular}{l|lll|lll|lr}
\hline
\textbf{Model} & \textbf{R2-P} & \textbf{R2-R} & \textbf{R2-F1} & \textbf{RL-P} & \textbf{RL-R} & \textbf{RL-F1}  & \textbf{ME} & \textbf{BLEU (\%)} \\ \hline
\hline
\multicolumn{9}{|c|}{\textbf{General-domain Large Language models}} 
\\ \hline

chatgpt-tldr  & \textbf{\textcolor{blue}{0.2391}} & 0.1428 & 0.1729 & \textbf{\textcolor{blue}{0.2956*}} & 0.1785 & 0.2149 & 0.1634 & 7.39  \\ \hline

chatgpt-summ  & 0.1964  & 0.1731 & 0.1818 & 0.2361 & \textbf{\textcolor{blue}{0.2087}} & 0.2188 & \textbf{\textcolor{blue}{0.1962}} & 10.82  \\ \hline

davinci-tldr  & 0.2338 & 0.1255 & 0.1568 & 0.2846 & 0.1529 & 0.1901 & 0.1412 & 6.82  \\ \hline
davinci-summ  & 0.2202 & \textbf{\textcolor{blue}{0.1795}} & \textbf{\textcolor{blue}{0.1954}} & 0.2513 & 0.2058 & \textbf{\textcolor{blue}{0.2234}} & 0.1917 & \textbf{\textcolor{blue}{11.41}}  \\ \hline

\multicolumn{9}{|c|}{\textbf{Legal domain-specific abstractive models}} \\ \hline

LegPegasus & 0.1964 & 0.1203 & 0.1335 & 0.2639 & 0.1544 & 0.1724 & 0.1943 & 13.14  \\ \hline

LegPegasus-IN & \textbf{\textcolor{blue}{0.2644}} & 0.2430 & 0.2516 & \textbf{\textcolor{blue}{0.2818*}} & 0.2620 & 0.2698 & 0.1967 & 18.66  \\ \hline

LegLED & 0.1115 & 0.1072 & 0.1085 & 0.1509 & 0.1468 & 0.1477 & 0.1424 & 8.43  \\ \hline

LegLED-IN & 0.2608 & \textbf{\textcolor{blue}{0.2531}} & \textbf{\textcolor{blue}{0.2550}} & 0.2769 & \textbf{\textcolor{blue}{0.2691*}} & \textbf{\textcolor{blue}{0.2711*}} & \textbf{\textcolor{blue}{0.2261}} & \textbf{\textcolor{blue}{19.81}}  \\ \hline

\multicolumn{9}{|c|}{\textbf{Extractive models}} 
\\ \hline
CaseSummarizer & \textbf{\textcolor{blue}{0.2512}} & \textbf{\textcolor{blue}{0.2269}} & \textbf{\textcolor{blue}{0.2381}} & \textbf{\textcolor{blue}{0.2316}} & \textbf{\textcolor{blue}{0.2085}} & \textbf{\textcolor{blue}{0.2191}} & 0.1941 & 15.46 \\ \hline


SummaRunner/RNN\_RNN & 0.2276 & 0.2103 & 0.2180 & 0.1983 & 0.1825 & 0.1893 & \textbf{\textcolor{blue}{0.2038}} & 17.58 \\ \hline

BertSum & 0.2474 & 0.2177 & 0.2311 & 0.2243 & 0.1953 & 0.2082 & 0.2037 & \textbf{\textcolor{blue}{18.16}} \\ \hline

\end{tabular}
\caption{Performance of summarization models from three families, that we have applied in this work. All metric values are averaged over the 100 documents in the IN-Abs test set. The metrics respectively are Rouge-2 precision, Rouge-2 recall, Rouge-2 F1 score, Rouge-L precision, Rouge-L recall, Rouge-L F1 score, METEOR and BLEU scores. The best value for every metric, for every family of summarization models, is shown in blue-bold.
Entries with an asterisk (*) indicate a value that is statistically significantly higher (by the Student T-test at 95\% confidence interval) than the best value achieved by an extractive summarisation model (the value shown in blue-bold) for the same metric.}
\label{tab:match-metrics}
\end{table*}

\subsubsection{Comparative results}

Table~\ref{tab:match-metrics} shows the performance of all the  summarization models (across the three families) that we have applied in this work, over the IN-Abs dataset. 
The best value for every metric in every family of summarization models is shown in blue-colored and boldfaced font.

We observe that out of the three families of summarization models, the legal domain-specific abstractive models achieve the best metric scores (better than both LLMs and extractive models). 
Extractive models achieve better scores than the general-domain LLMs for most of the metrics (ROUGE-2 scores, METEOR, BLEU), though the general-domain LLMs achieve slightly higher ROUGE-L scores.
We perform Student T-test at 95\% confidence interval to check if the best-performing abstractive model / LLM is performing statistically significantly better than the best-performing extractive model (individually for each metric). 
We see the improvements over the best extractive model are statistically significant only for the ROUGE-L metrics. The entries marked with an asterisk in Table~\ref{tab:match-metrics} indicate the values that are statistically significantly higher than the best value achieved by an extractive model for the same metric.

Out of the domain-specific abstractive models, LegPegasus-IN and LegLED-IN performed the best. The improvements in their performance over LegPegasus and LegLED show the benefits of in-domain finetuning (as stated in Section~\ref{sec:methods}, LegPegasus and LegLED are finetuned over US legal documents, but LegPegasus-IN and LegLED-IN are additionally finetuned on Indian legal documents similar to the IN-Abs test set). 

Though the LLMs (chatgpt and davinci)  achieve lower metric values than the best-performing abstractive and extractive models, 
their performance is creditable -- even though the LLMs have not been specifically trained over any legal dataset, they perform better than some of the extractive and abstractive models that are trained over legal data, at least according to certain metrics. 
For instance, davinci-summ achieves higher ROUGE-L F1 score than LegPegasus, LegLED and all the extractive models. 
Among the two variations of the LLMs, the `summ' variations achieve a little better scores than the `tldr' variations as per most metrics.




\subsection{Consistency of summaries}
\label{sub:summary-consistency}

We now check how consistent model-generated summaries are with the original documents. This check is important particularly for abstractive summarization models and LLMs which are known to hallucinate in text generation. 
We first describe the metrics, and then discuss comparative results.

\subsubsection{Metrics}

The following metrics compare the model-generated summary with the original document and estimate how consistent the summary is with the document. All these metrics give a score in the range $[0,1]$; the higher the score, the more consistent is the summary.



\textbf{(1) SummaC} -- This metric~\cite{laban2022summac} is based on Natural Language Inferencing (NLI) which is a task in Natural Language Processing that involves determining the relationship between two sentences.
One of the sentences is considered as a `hypothesis' and the other sentence is considered as a `premise'. NLI is the task of determining whether the given hypothesis logically follows from the premise. Typically, a NLI model will give a score representing how likely the hypothesis sentence is to logically follow from the premise sentence.

Given a (document, summary) pair, SummaC segments both the document and the summary into sentence units, and then leverages NLI models to effectively detect inconsistencies in the summary with respect to the document. 
In simple terms, NLI scores are computed for each sentence in the (model-generated) summary, to estimate the likelihood that this sentence logically follows from some sentence in the original document.  
Lower NLI scores for a particular sentence $s$ in the summary implies a higher mismatch between this sentence and the sentences in the original document, thus indicating a higher likelihood that this sentence $s$ contains hallucinated information. 
The NLI scores obtained by different sentences in the summary are then combined to give a single SummaC score for the given (document, summary) pair.
Thus, a higher SummaC score for a summary indicates that the summary is more consistent with respect to the original legal document (more details can be found in~\cite{laban2022summac}).

\textbf{(2) NumPrec} -- Numbers are an important part of a legal case judgement, because there are important numbers like dates, statute identifiers (e.g., Act and Section numbers), monetary values, terms of punishment, etc. 
It is important that these numbers are faithfully represented in the summary.
The NumPrec metric measures what fraction of the numbers present in the model-generated summary is also present in the source document. The numbers are identified using the standard Python library. 

\textbf{(3) NEPrec} --  Named Entities (NEs) are also very important in a legal case judgement. If entities like persons, organizations, etc. get changed in the summary, then not only will significant information be lost, but also the summary may become misleading. 
To detect the amount of inconsistency in a summary in terms of named entities, we calculate the metric called NEPrec that measures what fraction of the Named Entities present in the model-generated summary is also present in the source document. In this work, we detect Named Entities (from both the original document and the summaries) using the standard Spacy Toolkit available at \url{https://spacy.io/api/entityrecognizer}.

\vspace{2mm}
\noindent Note that the NumPrec and NEPrec metrics are dependent on the ability to detect numbers and named entities accurately. 
In particular, it is quite challenging to identify all types of named entities from Indian legal documents~\cite{kalamkar-etal-2022-named}. Hence the metric values are dependent on the accuracy of the Spacy toolkit used for this purpose. 


\subsubsection{Comparative results}

\begin{table}[tb]
\centering
\small
\begin{tabular}{llll}
\hline
\textbf{Model} & \textbf{SummaC} & \textbf{NEPrec} & \textbf{NumPrec} \\ \hline
\hline

\multicolumn{4}{|c|}{\textbf{General-domain Large Language models}} 
\\ \hline

chatgpt-tldr & 0.5719 & 0.8612 & 0.9498 \\ \hline

chatgpt-summ & 0.5762 & \textbf{\textcolor{blue}{0.9172}} & \textbf{\textcolor{blue}{0.9612}} \\ \hline

davinci-summ & \textbf{\textcolor{blue}{0.6356}} & 0.8959 & 0.9323  \\ \hline

davinci-tldr & 0.6080 & 0.8331 & 0.9123  \\ \hline

\multicolumn{4}{|c|}{\textbf{Legal domain-specific abstractive models}} \\ \hline

LegPegasus & 0.6333 & 0.8429 & 0.9483   \\ \hline

LegPegasus-IN & 0.7368 & \textbf{\textcolor{blue}{0.8542}} & \textbf{\textcolor{blue}{0.9952}}   \\ \hline

LegLED & 0.6563 & 0.7199 & 0.8192  \\ \hline


LegLED-IN & \textbf{\textcolor{blue}{0.8552}} & 0.8276 & 0.9769 \\ \hline


\end{tabular}

\caption{Consistency metrics of all abstractive methods and LLMs that we have applied in this work. All metric values are averaged over 100 documents in the IN-Abs dataset. 
The best value for every metric for each family of summarization models is highlighted.
}
\label{tab:consistency-metrics}
\end{table}

Table~\ref{tab:consistency-metrics} shows the performance of the LLM and abstractive summarization that we have applied in this work, over the IN-Abs dataset. All metric values are averaged over 100 documents. Note that it is meaningless to compute the metrics for extractive methods, since all the three metrics will be 1.0 by definition for any extractive method.

We now see some potential consistency issues with the LLMs and abstractive models. The SummaC scores for the LLMs are in the range $[0.5, 0.65]$ which show relatively lower consistency compared to the domain-specific abstractive models.
The NEPrec and NumPrec scores are higher, often higher than $0.9$; still these values indicate presence of some inconsistent / hallucinated named entities and numbers in the abstractive summaries. 

Among the domain-specific abstractive models, LegPegasus and LegLED have got relatively low scores (especially LegLED) which indicates substantial presence of hallucinated content in their summaries. 
LegPegasus-IN and LegLED-IN have consistently got higher scores (across all metrics) than the LegPegasus and LegLED models, which again shows the benefits of domain-specific finetuning. 


\subsection{Takeaways from this section}

The analyses in this section allows us to compare between extractive and abstractive summarization models, both trained over Indian legal documents. 
We see the abstractive models perform better than the extractive models according to standard metrics such as ROUGE, METEOR and BLEU (Table~\ref{tab:match-metrics}). Also the supervised models perform better than LLMs such as Davinci and ChatGPT.

However, abstractive models seem to have problems with consistency (Table~\ref{tab:consistency-metrics}). 
Some of the named entities / parts of the summary may be inconsistent with the original document. 
We look for the presence of such inconsistencies in the next section.

\begin{table*}[tb]
\centering
\small
\begin{tabular}{|p{0.01\textwidth}|p{0.12\textwidth}|p{0.33\textwidth}|p{0.45\textwidth}|}
\hline
\textbf{id} & \textbf{Model} & \textbf{Extract from summary showing error} & \textbf{Explanation of error}  \\ \hline

\hline

1
&
davinci-summ
&
The \textcolor{red}{language used, Deoria}, praying that the proceedings before the Nyaya Panchayat and its order dated December 25, 1963, be quashed ...
& 
As per the source document, `Deoria' is the name of a place, not the name of a language. So the sentence in the summary is meaningless. \\ \hline

2
&
chatgpt-summ
&
\textcolor{red}{The appellants, consisting of R Chari, M K Ramamurthi, Vineet Kumar, and Shyamala Pappu,} were found guilty of contempt of court and each sentenced ...
&
The names mentioned are actually that of the lawyers who represented the appellants, not the appellants themselves. The source document states ``A. S. R. Chari, M. K. Ramamurthi, Vineet Kumar and Shyamala Pappu, \textbf{for the} appellants''. The summarization model has mistakenly thought these names to be of the appellants themselves. \\ \hline

3
&
chatgpt-tldr
&
Mahabir filed an application under sections 4 and 5 of \textcolor{red}{theThe case} involves allegations of contempt of court
&
Incomplete sentence, where the name of the statute (Act) has been omitted in the summary. The most similar sentence in the main document is ``On May 21, 1964, Mahabir filed an application under ss. 4 and 5 of the Contempt of Courts Act, 1952, ...''
\\ \hline

4
&
LegLED
&
... violating the \textcolor{red}{antifraud provisions of Section 17(a) of the Securities Act of 1933, Section 10(b) of the Securities Exchange Act of 1934 and Rule 10b-5 thereunder,} ...
&
There is a lot of hallucination in this part of the summary. The phrases  ``Section 17(a) of the Securities Act of 1933'' and ``Section 10(b) of the Securities Exchange Act of 1934 and Rule 10b-5'' are all hallucinated. In particular, the Securities Act and Securities Exchange Act are Acts of the USA and are totally unrelated to the source document (which is a case in India). \\ \hline

5
&
LegLED
& 
On December 20, 1963, \textcolor{red}{the U.S. District Court for the Southern District of New York} entered a final judgment finding a judicial officer guilty of contempt of court for disobeying the order of the \textcolor{red}{U.S. District Court for the Southern District of New York.}
&
The “U.S. District Court for the Southern District of New York” that is stated in the summary has no relationship at all with this case (which is a case entirely argued in India) \\ \hline


\end{tabular}
\caption{Examples of errors in abstractive summaries generated by different models for the Indian Supreme Court judgement available at \url{indiankanoon.org/doc/1234444/}. The errors in the summaries have been marked in red. The last column explains the error.
}
\label{tab:errors-1234444}
\end{table*}


\begin{table*}[tb]
\centering
\small
\begin{tabular}{|p{0.01\textwidth}|p{0.10\textwidth}|p{0.35\textwidth}|p{0.42\textwidth}|}
\hline
\textbf{id} & \textbf{Model} & \textbf{Extract from summary showing error} & \textbf{Explanation of error}  \\ \hline
\hline

1
&
chatgpt-tldr 
&
The article examines three circumstances to determine whether the property in goods \textcolor{red}{passedThe} document discusses two separate legal cases related to the taxation ... 
& 
The first sentence is left incomplete and two sentences are merged. \\ \hline 

2
&
LegPegasus 
&
On September 27, 1960, the Supreme Court of India dismissed an appeal by \textcolor{red}{Daulatram Rameshwarlal and Daulatram Rameshwarlal J.M.} against the orders of the Bombay High Court ... 
&
The same name ``Dalutram Rameshwarlal'' is wrongly mentioned twice. There is no person called `Daulatram Rameshwarlal J. M.'' in the case. \\ \hline

3
&
LegPegasus 
&
The High Court held that the sale of castor oil \textcolor{red}{by M/s. Daulatram Rameshwarlal to M/s. Daulatram Rameshwarlal Ltd} was exempt from purchase tax under the provisions ... 
&
The same entity (M/s. Daulatram Rameshwarlal) is stated both as the seller and buyer, which is wrong. \\ \hline

4
&
LegPegasus 
&
The Court of Appeal held that it is the \textcolor{red}{duty of the buyer to obtain the necessary export licence}. The Court of Appeal held that \textcolor{red}{it was for the sellers to obtain the licence} and this view was approved by the House of Lords. 
&
The first line says getting the licence is the duty of the buyer, but the immediate next line says it is the duty of the seller – this is inconsistent. 

In the source document, the relevant part says that the ordinary rule in FOB contracts is that it is the duty of the buyer to obtain the export licence, but there was one special case where it was deemed to be the duty of the sellers. This meaning is lost in the summary. \\ \hline

5
&
LegLED 
& 
On September 27, \textcolor{red}{2019}, the \textcolor{red}{U.S. District Court for the Southern District of New York} entered a final judgment against \textcolor{red}{Daulatram Rameshwarlal, a firm} registered under the Indian Partnership Act, and \textcolor{red}{Daulatram Rameshwarlal, a registered dealer} under the Indian Partnership Act, for claiming exemption from Sales Tax in respect of sales of cotton ... 
&
The `U.S. District court of New York' is hallucinated (the original case was argued entirely in Indian courts). Also the year `2019' is hallucinated -- the original case is of 1961, so no event of 2019 could have been referred. 

Also, the summarization model did not understand that the same entity `Daulatram Rameshwarlal' is referred to both as a `firm' and a `registered dealer'; the model has assumed two separate entities. \\ \hline

6
&
LegPegasus-IN 
& 
The intention of the parties that in compliance with the requirements of cl.5(2) of the Exports (Control) \textcolor{red}{OrderThere} is no circumstance which would justify a conclusion that ...
&
The first sentence is left incomplete and two sentences are merged.\\ \hline


7
&
LegLED-IN 
& 
\textcolor{red}{The Court was right in holding that the Court was wrong in holding that} it was not necessary 
&
This sentence in the summary is meaningless. The source document is a case heard in the Supreme Court of India, and is an appeal against a decision pronounced by the Bombay High Court. Hence two courts are involved, but it is not clear from the summary which court is being referred to by which occurrence of the word `court'.
\\ \hline

\end{tabular}

\caption{Examples of errors in abstractive summaries generated by different models for the Indian Supreme Court judgement available at \url{https://indiankanoon.org/doc/27285/}. The errors in the summaries are marked in red, and explained in the last column.}
\label{tab:errors-27285}
\end{table*}

\section{Inconsistencies in abstractive summaries}

The analysis in Section~\ref{sub:summary-consistency} indicates that some parts of the summaries generated by abstractive models and LLMs may {\it not} be consistent with the original documents.
To understand what kind of inconsistencies are present in the summaries, we manually observed a large number of (document, summary) pairs from our dataset. 
In particular, we observed those sentences that obtained relatively low SummaC scores, and those sentences that contained numbers and named entities that could not be matched with the original documents (while computing NERPrec and NumPrec). We also observed the relevant parts in the main document to understand the errors/inconsistencies.

We found several different types of errors and inconsistency in the abstractive summaries. 
Table~\ref{tab:errors-1234444}, Table~\ref{tab:errors-27285}, Table~\ref{tab:errors-1722864} show some example errors/inconsistencies in the summaries generated by the abstractive models and LLMs for three specific Indian Supreme Court documents (which are mentioned in the table captions). 
The tables show the name of the model, an extract from the summary showing the error, and an explanation of the error.

We observed some common types of errors in most summaries generated by almost all abstractive models and LLMs, such as \textbf{two sentences being merged} (leaving the first sentence incomplete) -- for examples, see Table~\ref{tab:errors-1234444} error-3, Table~\ref{tab:errors-27285}, error-1 and Table~\ref{tab:errors-1722864} error-4. 
These errors mostly happen at the boundary of chunks. 

We also observed more serious errors such as \textbf{wrong numbers being generated in the summary}, which are not present in the original document. 
For instance, Table~\ref{tab:errors-27285} error-5 shows a wrong year being mentioned in the summary -- this table refers to a case heard in 1961; hence the year `2019' in the LegLED summary is clearly hallucinated.

\begin{table*}[tb]
\centering
\small
\begin{tabular}{|p{0.01\textwidth}|p{0.10\textwidth}|p{0.35\textwidth}|p{0.42\textwidth}|}
\hline
\textbf{id} & \textbf{Model} & \textbf{Extract from summary showing error} & \textbf{Explanation of error}  \\ \hline
\hline
1
&
LegLED 
& 
On March 31, 1965, the Honorable \textcolor{red}{M.K. Ramaswami} of the Madras High Court granted the SEC's request for an asset freeze and other emergency relief. 
&
The name of the judge in the source document is `V. Ramaswami' (and not `M.K. Ramaswami' as stated in the summary). Whereas, `M.K. Ramamurthi' is one of the lawyers representing the appellant. The summarization model has confused between the two names. \\ \hline

2
&
LegLED & 
The SEC's complaint, filed in the \textcolor{red}{U.S. District Court for the Southern District of Madras}, alleges that ... &
A wrong court has been mentioned. This is a case in India, hence “U.S. District Court” is hallucinated by the summarization model. \\ \hline

3
&
LegLED 
&
The phrase ``regulated by usage'' in \textcolor{red}{section 6(9) of the MadrasHereditary} succession is succession by the heir to the deceased under the law, the office must be transmitted to the successor according to some definite rules of descent which by their own force designate the person to succeed. 
&
The name of the Act has been left incomplete (actually, `The Madras Hindu Religious and Charitable Endowments Act, 1951') , and the word ``Madras'' has been merged with the next sentence. \\ \hline

4
&
LegPegasus-IN 
& 
The word \textcolor{red}{"successionIt} is true that the artificial definition of hereditary trustee in section 6(9) of the Act would include even such cases. 
& 
One sentence has been left incomplete and the word ``succession'' has been merged with the next sentence. Note that the sentence that has been left incomplete is an important sentence where the court explains its interpretation of the word ``succession'' in the context of this case. \\ \hline

\end{tabular}

\caption{\label{tab:errors-1722864}
Examples of errors in abstractive summaries generated by different summarization models for the Indian Supreme Court judgement available at \url{https://indiankanoon.org/doc/1722864/}. The errors in the summaries are marked in red, and explained in the last column.
}
\end{table*}

We noticed one strange type of error particularly in summaries generated by LegLED -- even when the models are summarizing Indian case judgements,  names of U.S. Courts and names of U.S. statutes come up in the summaries, which are not at all related to the input document. Examples of such hallucinations are shown in  Table~\ref{tab:errors-1234444}, error-4 and error-5, and Table~\ref{tab:errors-1722864} error-2. 
Such hallucinations are probably due to the fact that LegLED has been trained on US legal document-summary pairs, and the model has a tendency of generating US court / statute names that it has seen during training.
Importantly, we did \textit{not} observe this type of error in the LegLED-IN summaries, which shows that domain-specific fine-tuning can help to reduce hallucinations.
Also we did {\it not} observe this particular type of error in the summaries generated by the LLMs (ChatGPT or DaVinci).

There are also examples of \textbf{errors in named entities}, e.g., a case where LegLED confused the name of a judge with the name of a lawyer (Table~\ref{tab:errors-1722864} error-1) and a case where chatgpt-summ mistakenly thought the lawyers representing the appellants to be the appellants themselves (Table~\ref{tab:errors-1234444} error-2). 
Such errors are very difficult to detect by automatic methods, and can lead the summaries to be misleading.


\section{Concluding discussion}

We have tried a wide range of Large Language Models (e.g., Text-Davinci-003 and Turbo-Gpt-3.5) and domain-specific abstractive summarization models (e.g., Legal-LED, Legal-Pegasus) on a dataset of Indian Supreme Court case judgements, and calculated a wide range of metrics. 
Apart from the standard metrics of evaluation like ROUGE, METEOR, BLEU, we also calculate non-traditional metrics for evaluation of summary consistency like Numprec, NERprec and SummaC. 

We observe that the domain-specific fine-tuning improves the performance of abstractive models (LegPegasus-IN and LegLED-IN) in terms of both match with gold standard summary and consistency. 
LLMs such as Turbo-GPT-3.5 (ChatGPT) and Text-Davinci-003 also perform well in a zero-shot setting, considering they have not been trained specifically on legal documents. However, these LLMs also sometimes generate inconsistent text in summaries.  

In general, we see that the abstractive models often outperform the extractive models in terms of metrics such as ROUGE, METEOR and BLEU (Table~\ref{tab:match-metrics}). 
However, the abstractive models are fraught with issues like inconsistencies and hallucinations in the generated summaries. 
Some of the problems can be mitigated by \textit{domain-specific fine-tuning}; for instance, while LegLED often generates names of US courts/statutes while summarizing Indian documents, these errors are considerably lesser in LegLED-IN which is further fine-tuned on Indian legal data. 
Some of the errors can also be potentially detected and addressed by careful post-processing of the generated summaries.
However, some of the errors committed by abstractive models are subtle and much more difficult to detect automatically, e.g., confusing the names of appellants and the names of the lawyers representing the appellants (see the third example in Table~\ref{tab:errors-1234444}). To our knowledge, this is the first work to demonstrate examples of such complex errors in abstractive summaries of legal case judgments.

So, as expressed by the experiments reported in this paper, we conclude 
(1)~pre-trained abstractive summarization models and LLMs are not yet ready for fully automatic summarization in a complex domain such as Law; possibly a human-in-the-loop approach is more suitable where a legal expert can monitor the quality of the summaries generated by these methods, and 
(2)~better methods need to be designed to detect complex types of errors in abstractive summaries. 
In future, we plan to pursue these directions towards improving abstractive summarization in the legal domain.


\vspace{5mm}
\noindent \textbf{Acknowledgements:}
The authors acknowledge useful feedback and suggestions about the work from Jack Conrad (from Thomson Reuters Labs).
The research is partially supported by the TCG Centres for Research and Education in Science and Technology (CREST), India through a project titled ``Smart Legal Consultant: AI-based Legal Analytics''.



\end{document}